\documentclass[letterpaper, 10 pt, conference]{ieeeconf}  
\IEEEoverridecommandlockouts   
\usepackage[T1]{fontenc}
\usepackage[latin9]{inputenc}
\usepackage{color}
\usepackage{textcomp}
\usepackage{graphicx}
\usepackage{array,booktabs}
\usepackage{graphicx}
\usepackage{algorithm,algorithmicx,algpseudocode}
\usepackage{amsmath} 
\usepackage{amssymb}  
\usepackage{multirow}
\usepackage{epstopdf}
\usepackage{siunitx}
\usepackage{mathtools}
\usepackage{subfigure}
\usepackage{multicol}

\pdfmapfile{=fullembed.map}

\title{\LARGE \bf
	A Statistical Update of Grid Representations from Range Sensors
}

\author{Luis Rold\~ao$^{1,2}$, Raoul de Charette$^{1}$ and Anne Verroust-Blondet$^{1}$
	\thanks{$^{1}$The authors are with the Robotics and Intelligent Transportation Systems (RITS) Team, INRIA Paris, 2 Rue Simone Iff, 75012 France. {\tt\small \{raoul.de-charette, anne.verroust\}@inria.fr}.} \thanks{$^{2}$Author is with the R\&D Department of AKKA Technologies. 78280 Guyancourt, France. {\tt\small \{luis.roldao@akka.eu\}}.}
}

\begin{document}
	
\maketitle
\thispagestyle{empty}
\pagestyle{empty}

\begin{abstract}	
	In a wide range of robotic applications, being able to create a 3D model of the surrounding environment is a key feature for autonomous tasks. In this research report, we present a statistical model to perform 3D reconstructions of the environment from range sensors using an occupancy grid. To do so, we take into account all the available information obtained from the sensor, considering the distances traversed by the rays in each cell and seeking to reduce reconstruction errors caused by discretization. The approach has been validated qualitatively using the KITTI dataset.
\end{abstract}

\section{Introduction}\label{Introduction}

In robotics, an accurate 3D model of the world is fundamental for performing different tasks \cite{Thrun02roboticmapping:}. While continuous representations may be used \cite{Curless1999, Newcombe6162880}, discrete grid representations are preferred in order to reduce memory and computational complexity \cite{Moravec-1985-15232}. In this case, each grid cell represents the occupancy state of a portion of the environment. Most popular methods aim to represent this state probabilistically \cite{Thrun:2005:PR:1121596}.

Initially implemented in 2D, grids were further extended in several works. In \cite{30717Bares} and \cite{100111Herbert}, an additional height value is assigned to each cell. Moreover, several surfaces can be also stored \cite{Triebel4058725}. More recently, the wide use of 3D range sensors and the increase of computational capabilities have boosted the popularity of volumetric grids with cubical voxel cells.

By definition, a discretized representation inhibits a completely accurate reconstruction. Therefore, grid models are unable to create a perfect model of the surroundings. Discretization inaccuracies can be reduced by using a smaller cell size \cite{Milstein2008og,DIA201713841}, but this increases the computation and memory needs. 

\begin{figure}[t]
	\begin{center}
		\includegraphics[width=0.48\textwidth]{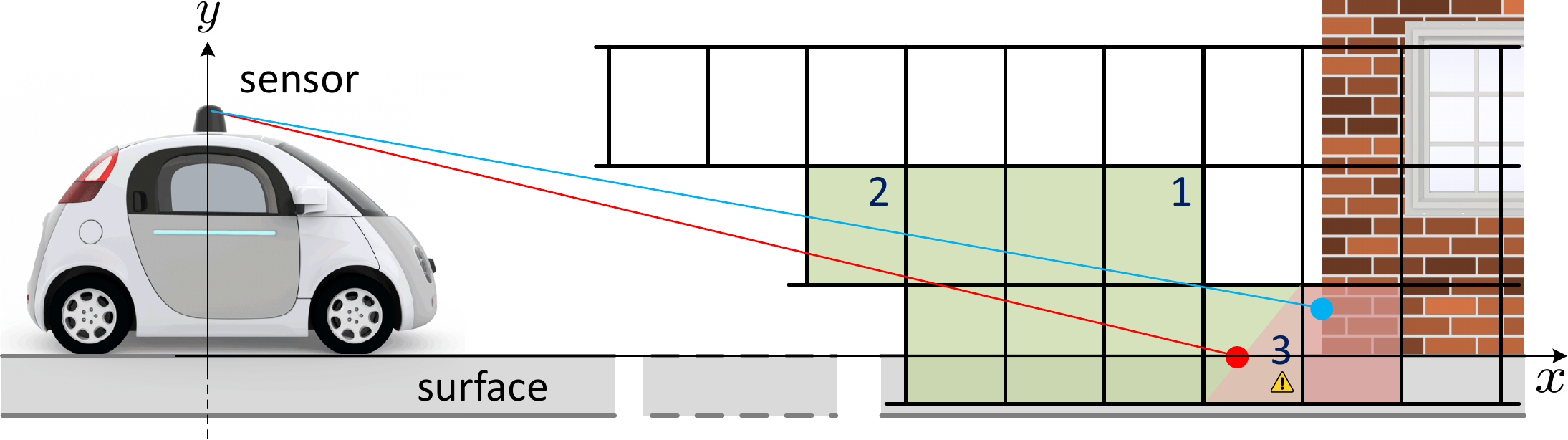}
		\caption{Most occupancy grid methods consider that the complete state of a cell within a scan can be updated from a single measurement. Under this assumption, a slightly traversed cell (as 1 in the Figure) will have the same occupancy probability than cells traversed by several rays (see cell 2). Moreover, partially occupied cells as 3 will have the same probability than a completely occupied cell if only the impact measurement is considered.}
		\label{fig:DiscretizationError}
		\vspace{-0.3cm}	
	\end{center}
\end{figure}

Recursive structures such as quadtrees  and octrees \cite{Hornung2013} are commonly used to face this problem, leading to a $\mathcal{O}(\log{}n)$ complexity for insertions and queries. However, while this allows a finer map representation (i.e. smaller cell-size) at smaller memory cost, discretization still inhibits the possibility of representing cells partially occupied. To tackle this problem, most methods give higher priority to some measurements and ignore repeated observations as illustrated in Fig \ref{fig:DiscretizationError}.

We propose to insert all the single measurements obtained from a lidar sensor in a statistical way, exploiting all the available information about the ray path and considering the traveled distances of the rays at each cell (traversed and impacted ones) in order to perform a more accurate map update. Details of our method are provided and a comparison is done against the widely used OctoMap \cite{Hornung2013}.

\section{Voxelized Representation}\label{VoxelizedRepresentation}

In volumetric grids, the occupancy state of each voxel is recursively updated by performing a ray-casting operation that tracks each ray from the sensor origin to the impact point. In the literature, voxels that have been traversed by a ray ---\textit{misses}--- can be considered as free, while voxels where an impact occurs ---\textit{hits}--- are considered as occupied. 

In this line of work, most of the approaches insert the different sensor measurements acquired over time by applying a static state binary Bayes filter as introduced in \cite{Elfes30720}, where the estimated probability of a particular voxel $v$ being occupied $P(v|z_{1:t})$ is defined as:

\vspace{-0.4cm}
\begin{multline} \label{eq:BayesFilter}
P(v | z_{1:t}) = \\ 
\left[1 + \frac{1-P(v | z_{1:t-1})}{P(v | z_{1:t-1})} \frac{1-P(v | z_{t})}{P(v | z_{t})} \frac{P(v)}{1-P(v)} \right]^{-1} 
\end{multline}
\vspace{0.06cm}

\noindent where $z_{1:t-1}$ corresponds to all previous measurements acquired; $z_{t}$ expresses the current observation at time $t$, and $P(v)$ represents the prior knowledge about the map. Commonly, defined $P_{free}$ and $P_{occ}$ probability values are assigned to $P(v|z_{t})$ for misses and hits, respectively. If the initial state of cells is unknown, then $P(v) = 0.5\,$. In order to output a binary occupancy map (i.e. occupied or free cells), probabilities of the cells are usually thresholded at 0.5.   

\section{Approach}\label{Approach}

In the literature, it is usually considered that within a single scan, the state of each cell is binary (free or occupied). Hence, a cell is set occupied if at least one impact occurred within, and free if it has been traversed by any ray. The problem of such approach is that the complete state of the cell is updated from a single partial observation, neglecting the contribution of multiple measurements and their validity. As an example, a cell will be updated as free or occupied with the same probability regardless of how the ray traversed it or where the impact occurred, respectively. Moreover, a cell traversed by several rays will have the same occupancy probability than another cell traversed only once. This is shown in Fig. \ref{fig:DiscretizationError}.  

Conversely, our method updates the occupancy probability of each cell by considering the ray path information, and the density of observations that can be obtained at such cell, which depends on its distance from the sensor.

\subsection{Weight Measurement Probability}\label{Weight_funct}

3D sensors mounted on a mobile platform produce a sparse and uneven distribution of sensor data, where the density of measurements is high at close ranges and decays rapidly with the distance. Consequently, closer cells will have a considerably higher number of observations than farther ones. To account for this, we propose to weight the occupancy update of each cell according to the density of observations $\rho(d)$ that the sensor can measure from a cell at distance $d$.

To do so, we have calculated a function $\rho(d)$ that physically models the approximate number of rays that might traverse a voxel of a particular size $\omega$ at a given distance $d$, depending on the vertical and horizontal angular resolutions of the sensor, $\varphi_s$ and $\theta_s$, respectively. This function has been estimated by doing approximations in order to ease the complexity of the calculation, and experimentally validated by comparing the result against synthetic data (We refer the reader to the Appendix for more details).

\begin{figure}[h]
	\centering
	\subfigure[]{\label{fig:FunctionsWeightAndModel_a}\includegraphics[width=38mm]{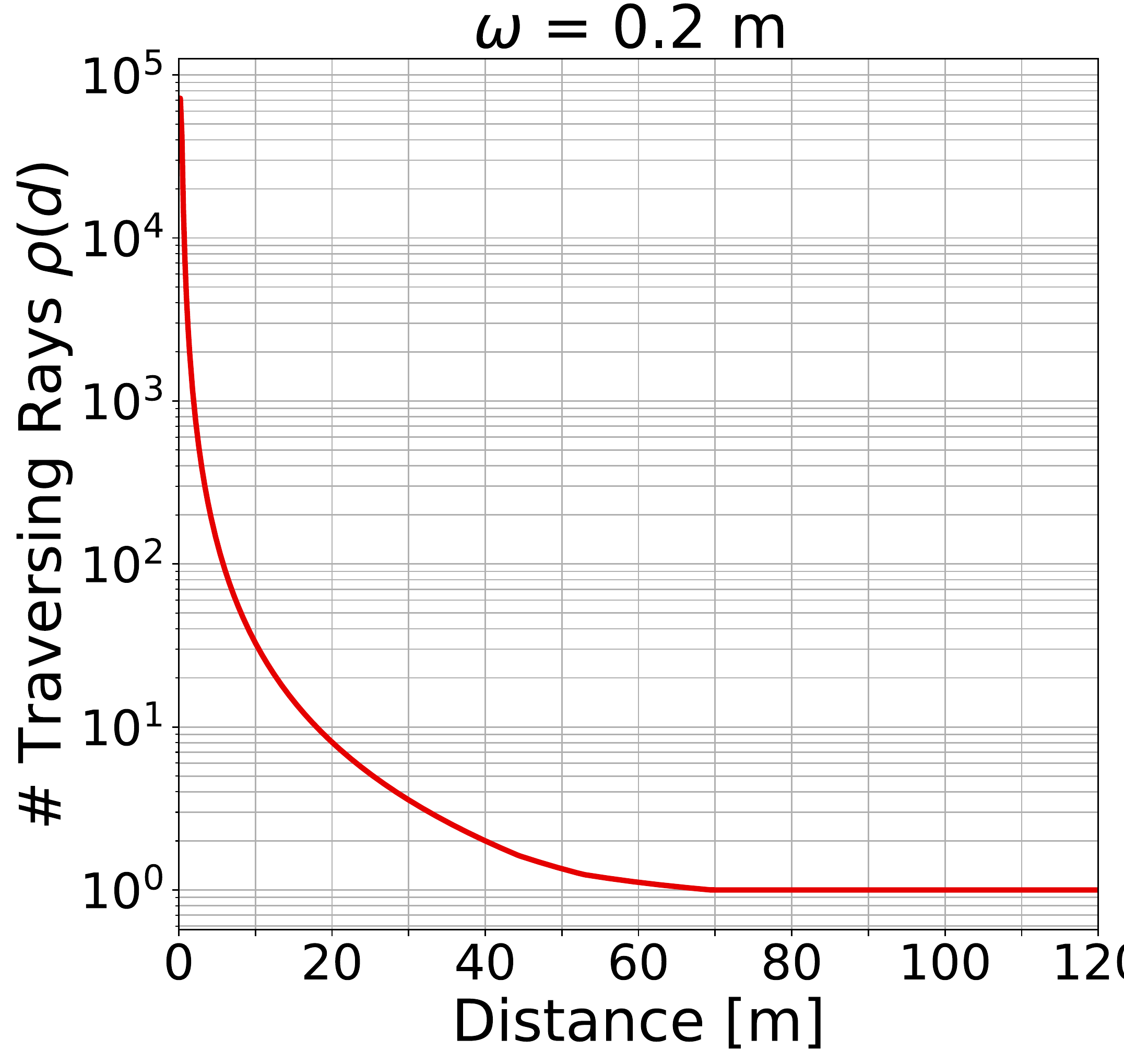}}
	\subfigure[]{\label{fig:FunctionsWeightAndModel_b}\includegraphics[width=38mm]{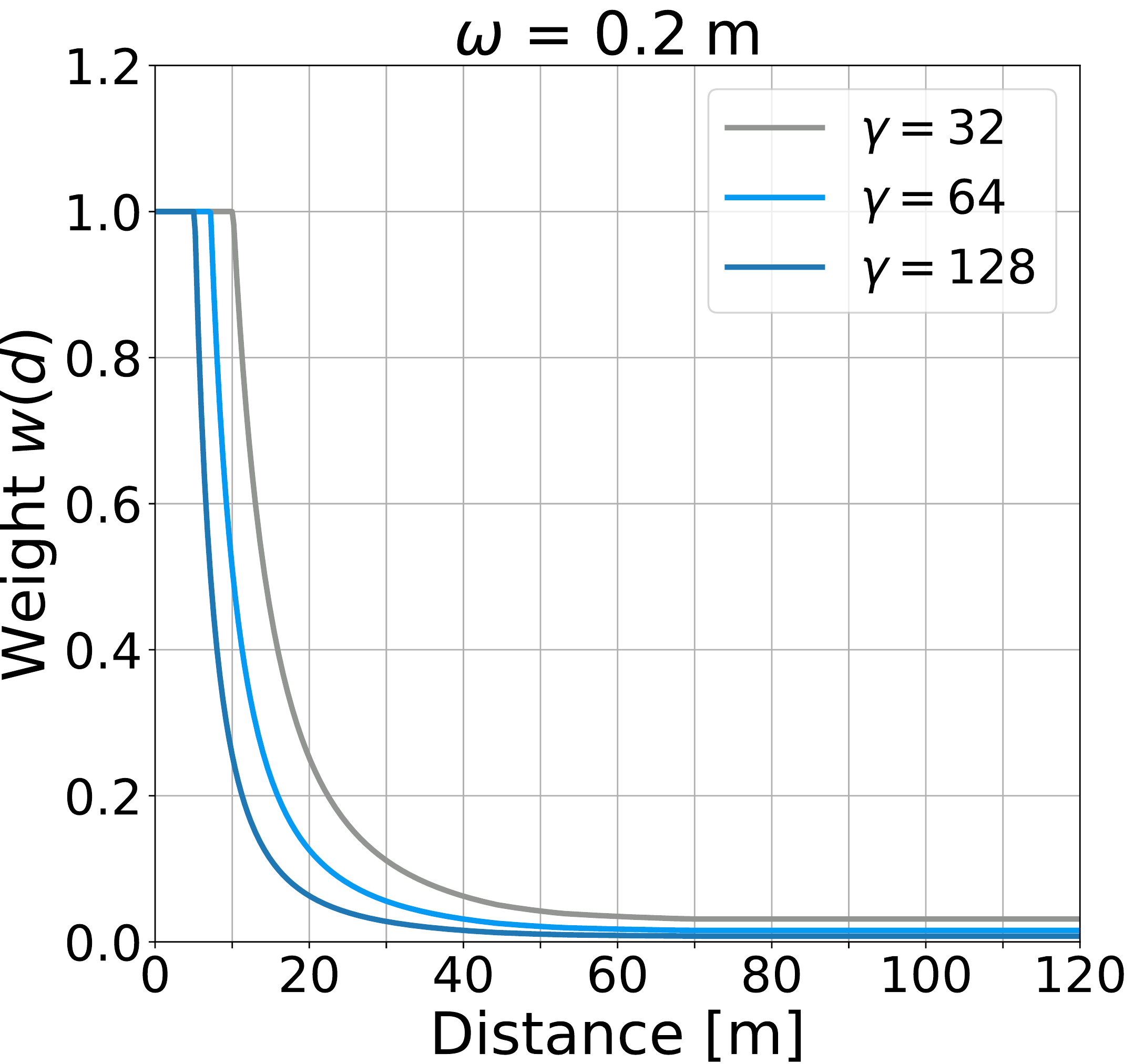}}
	\vspace{-0.1cm}
	\caption{(a) An example of the modeling function $\rho(d)$ at a $\omega = 0.2m$ voxel size. Number of observations decrease exponentially with the distance. (b) Example of weight function $w(d)$ with different $\gamma$ set at a voxel size $\omega = 0.2m$. For both cases we set $\theta_s = \ang{0.16}$ and $\varphi_s = \ang{0.4}$.}
\end{figure}

The weighting function $w(d)$ is then obtained by scaling $\rho(d)$ according to $\gamma$ and clamping it to 1. $\gamma$ will affect the drop-off of the function and the distance at which the measurements will start being weighted, as it can be seen on Fig. \ref{fig:FunctionsWeightAndModel_b}. The weighting function is then defined as:

\begin{equation}\label{eq:TheWeightingFunction}
w(d) = \min\left(1, \frac{\rho(d)}{\gamma}\right)
\end{equation} 

\subsection{Occupancy Probability From Traversability}

As opposed to most approaches in the literature, we propose to consider the ray path information (i.e. distance traveled by the ray) within each voxel in order to weight its occupancy probability. This is based on the fact that rays are only partial observations, and the information completeness depends on how all rays traverse each cell. A similar approach has been proposed in \cite{7857032Schaefer}, but only traveled distance within completely traversed cells is considered. 

In our case, we also consider the distance traversed within impacted voxels (from the voxel's border to the coordinates of impact) to assign the respective occupancy probability. Just as in \cite{Hornung2013} we define fixed $P_{free}$ and $P_{occ}$ observation probabilities that are below and above 0.5 respectively, and extend their sensor model by linearly weighting such probabilities according to the ray path information within the cell.

\vspace{-0.1cm}	
\begin{figure}[H]
	\begin{center}
		\includegraphics[width=0.48\textwidth]{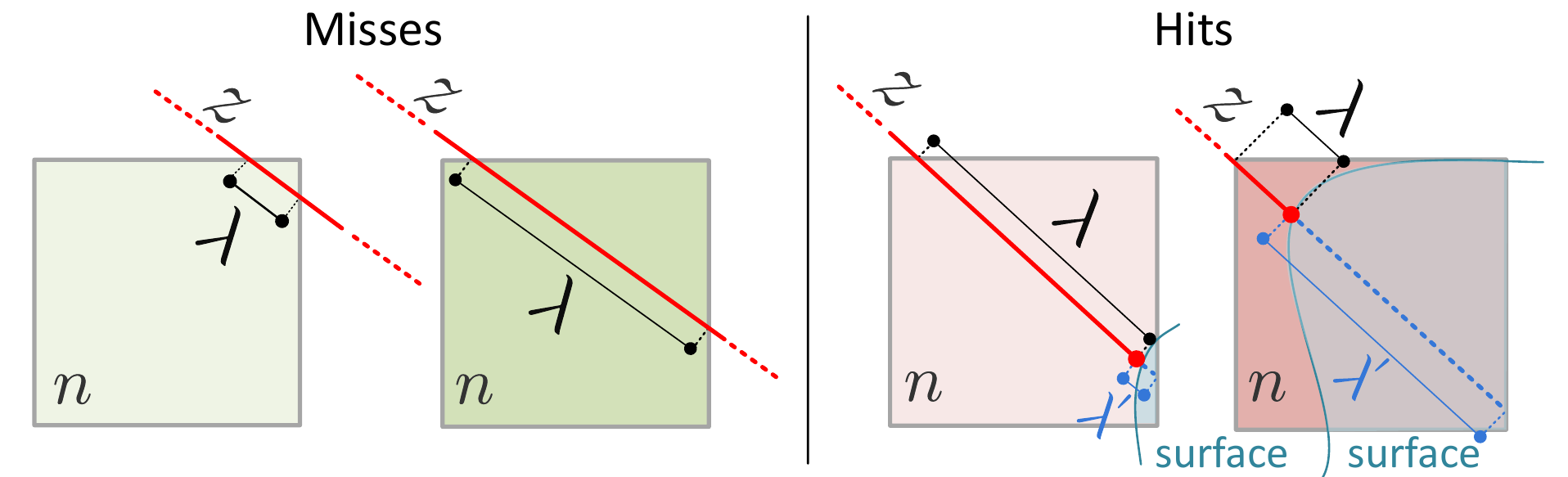}
		\vspace{-0.7cm}	
		\caption{The Occupancy Probability $P(n | z)$ is modulated according to the traversed distance $\lambda$ of the measured ray $z$ within the cell $n$.}.
		\label{fig:ProbExpl2}		
	\end{center}
\end{figure}
\vspace{-0.4cm}

Our intuition is that as higher is the distance that a ray has traversed within a particular cell $\lambda$, lower should be its occupancy probability (see Fig. \ref{fig:ProbExpl2}). For the traversed cells we modulate this probability by comparing $\lambda$ with the diagonal length of the cell ($\sqrt{3}\omega$), which corresponds to the maximum distance that a ray can traverse. In the case of impacted cells, the comparison is done with the length that the ray would have traversed within the cell if it wouldn't encounter any obstacle $(\lambda + \lambda')$, which corresponds to the maximum possible traversed distance given the current observation. 

\begin{figure*}[t]
	\begin{center}
		\centering
		\subfigure[OctoMap \cite{Hornung2013}]{\label{fig:MethodComparisonA}\includegraphics[width=57mm]{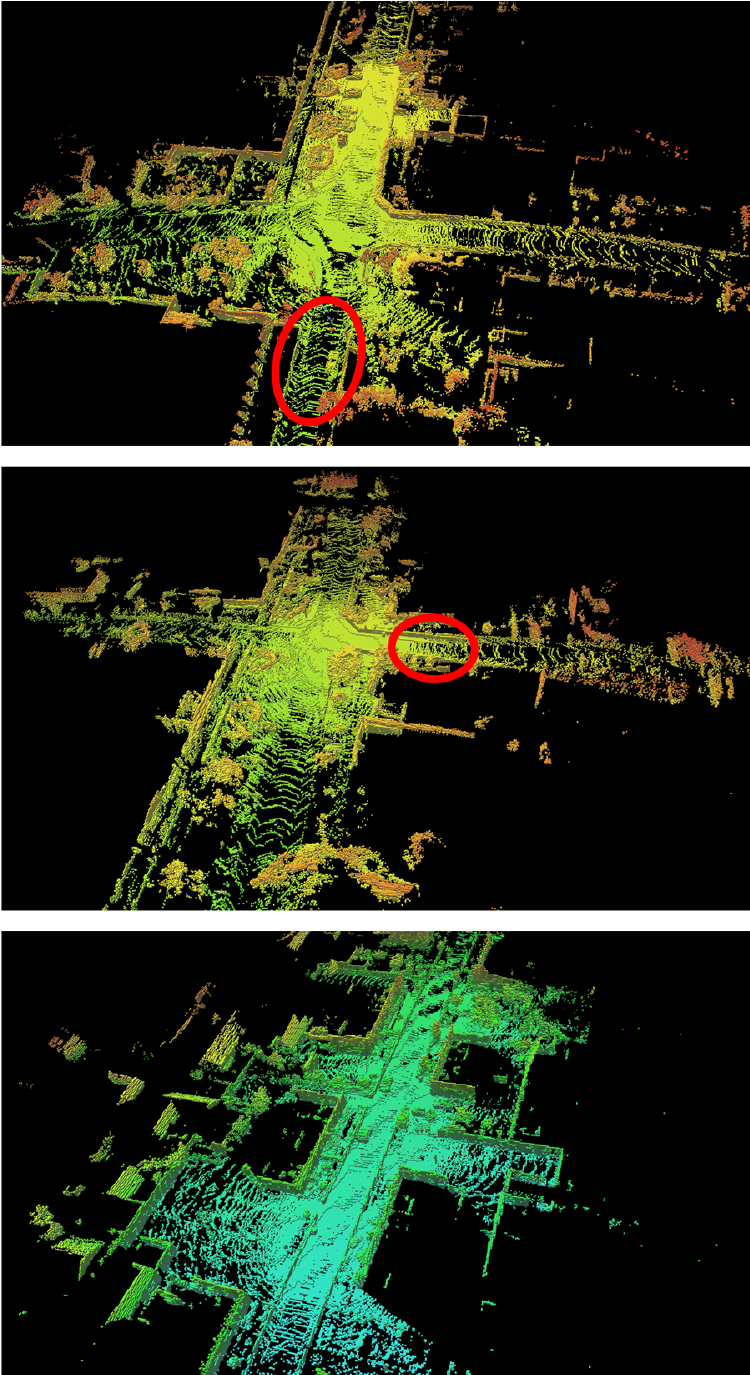}}
		\subfigure[Our approach - Method 1]{\label{fig:MethodComparisonB}\includegraphics[width=57mm]{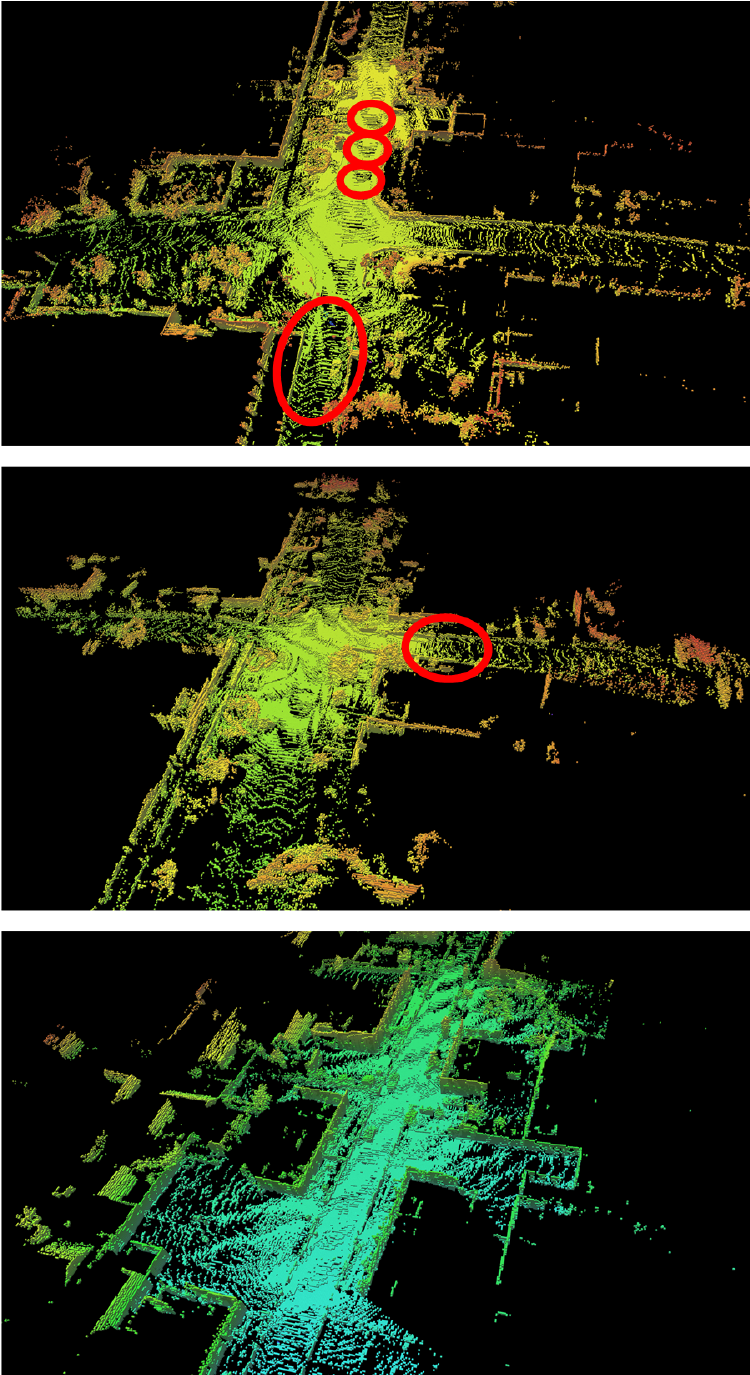}}
		\subfigure[Our approach - Method 2]{\label{fig:MethodComparisonC}\includegraphics[width=57mm]{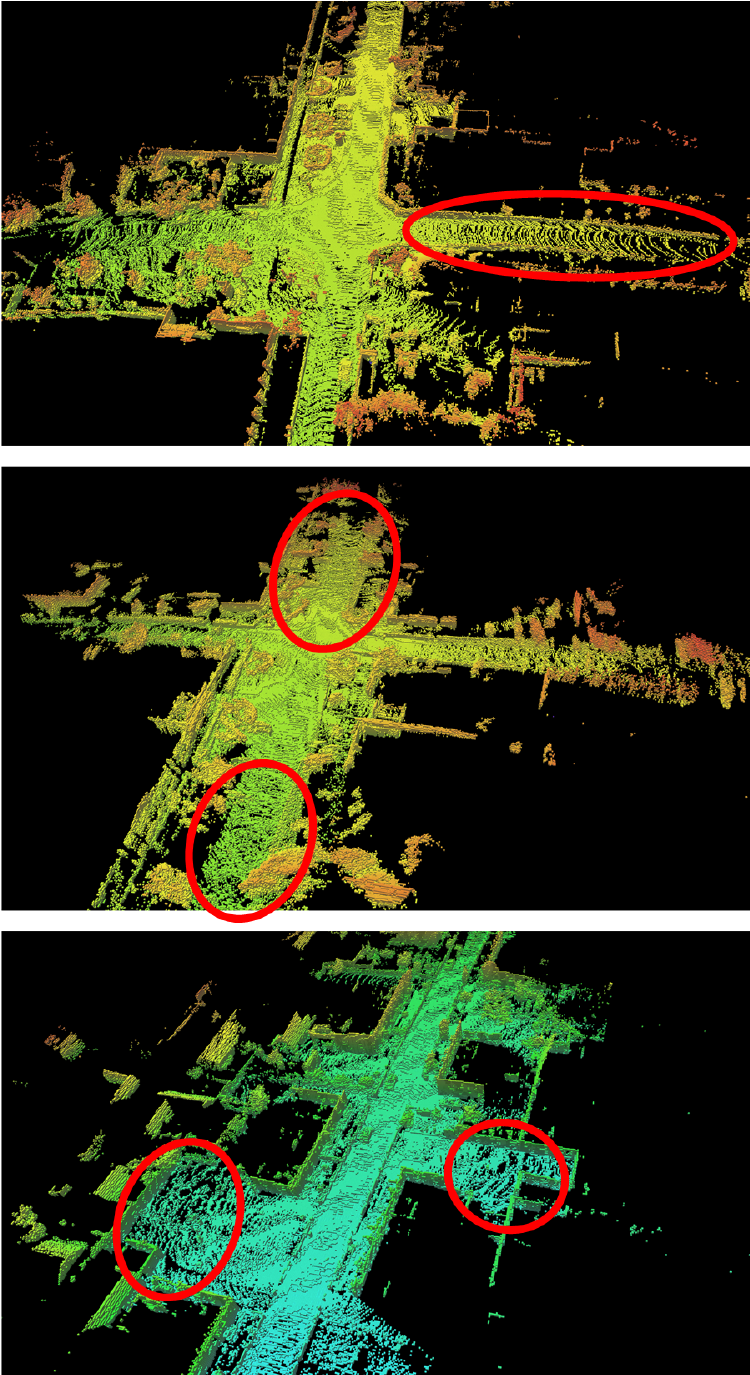}}
		\vspace{-0.1cm}
		\caption{Method comparison columnwise. Each row represent a subset of 60 frames integrated together into an octree map using odometry and RTK-GPS data as pose of the vehicle. Column (a) presents the approach of \cite{Hornung2013}, where not all the measurements are considered and traversal observations are ignored if impacts have occurred in a particular cell. Column (b) presents our first proposed method, considering all measurements in a single scan, the traversed distances within cells, and the sensor-voxel distance weighting function for both misses and hits. Column (c) presents results of our second method by applying the weighting function to the traversal observations only. The goal is to obtain a dense reconstruction with almost no free voxels in the area corresponding to the ground.}
		\vspace{-0.3cm}
		\label{fig:MethodComparison}
	\end{center}
\end{figure*}

\subsection{Occupancy Update}
 
In order to update the occupancy probability of a voxel $v$, we present two different methods. In the first method we account for all the information obtained from the complete set of measurements in a single scan (i.e. all rays traversing and/or impacting the voxel), weighting all the observations according to the rays traversed distance and the sensor density. The contribution of each measurement $m_i$ in a single scan at time $t$ is assigned as defined in the following equation: 

\small
\begin{multline}\label{eq:SensorModelDev}
P_{M1}(v | z_{t, m_i}) =
\begin{dcases}
0.5 + \left[\frac{(P_{occ}-0.5) \, \lambda'_i}{\lambda_i+\lambda'_i}\right] w(d)
&
\text{for a hit}
\\
0.5 - \left[\frac{(0.5-P_{free})\lambda_i}{\sqrt{3}\omega}\right] w(d)
&
\text{for a miss}
\end{dcases}
\end{multline}
\normalsize

Alternatively, in the second proposed method, only misses measurements are weighted by $w(d)$, giving a higher influence to hits measurements. This is commonly done in  literature \cite{Hornung2013, 7857032Schaefer}. Each measurement is then updated by:

\small
\begin{multline}\label{eq:SensorModelDev}
P_{M2}(v | z_{t, m_i}) =
\begin{dcases}
0.5 + \left[\frac{(P_{occ}-0.5) \, \lambda'_i}{\lambda_i+\lambda'_i}\right]
&
\text{for a hit}
\\
0.5 - \left[\frac{(0.5-P_{free})\lambda_i}{\sqrt{3}\omega}\right] w(d)
&
\text{for a miss}
\end{dcases}
\end{multline}
\normalsize

\section{Experimental Results}\label{Validation}

The proposed methods were implemented in C++ using the OctoMap library \cite{Hornung2013} and evaluated on one of the residential subsets ($2011\char`_09\char`_30\char`_$drive$\char`_0018$) of the KITTI dataset \cite{Geiger:2013:VMR:2528331.2528333}, acquired from a roof-mounted HDL-64E lidar sensor. 

We used 3 subsequences of 60 frames to evaluate qualitatively the results of our approaches since no ground truth is available. Our results were compared with OctoMap \cite{Hornung2013}, where only one observation per scan is considered and ray path information is not accounted. The provided RTK-GPS data was used for frame-to-frame registration. All the maps were created using octrees of $ \omega = 0.2m $ leaf voxel size since it represented the best trade-off between computational time and accuracy. 

As in \cite{Hornung2013}, the occupancy probabilities of all the cells were initialized with a uniform prior of $P(n_i) = 0.5$. Equally, values of $P_{free} = 0.4$ and $P_{occ} = 0.7$ for the misses and hits respectively were used. Clamping thresholds of $P_{min}=0.12$ and $P_{max}=0.97$ were set in order to limit the probability values that can be assigned to the volumes to update occupancy changes in the environment in a fast manner. For our weighting function, we used a value of $\gamma = 32$, the vertical and horizontal angular resolutions of the weight function $\varphi_s$ and $\theta_s$ were set as $\ang{0.4}$ and $\ang{0.16}$, corresponding to the elevation and azimuth values of the velodyne HDL-64E lidar sensor.

Qualitative results are presented in Fig. \ref{fig:MethodComparison}, where each row corresponds to a 60 frames subset inserted into an octree map. In Fig. \ref{fig:MethodComparisonA} reconstruction results using OctoMap \cite{Hornung2013} are shown, notice how an important proportion of voxels are incorrectly classified as free, specially in the ground region. In Fig. \ref{fig:MethodComparisonB} our first approach is presented, applying the weighting function $w(d)$ to both misses and hits measurements. Fig. \ref{fig:MethodComparisonC} presents the results of our second approach by applying $w(d)$ to misses only. Frames used for this tests did not contain any dynamic objects. Ideal results consist in a dense reconstruction with most of the environment filled and no holes in the ground.

Although our first method considers both, the traversed distances of all rays within the cells, and the weight of the measurements according to its distance from the sensor, no significant differences can be appreciated between Fig. \ref{fig:MethodComparisonA} and Fig. \ref{fig:MethodComparisonB} columns. In both cases, discretization inaccuracies can be observed, particularly in the area corresponding to the ground (circled in the figure). In our case, a partially occupied voxel can be represented as free if the influence of the traversed rays is stronger than the one of impacts.

In Fig. \ref{fig:MethodComparisonC}, a better representation can be seen, with fewer holes in the ground and a more dense reconstruction, specially on the circled areas. This is due to the weighting of the misses observations only, which reduces the influence of rays traversing cells that are far from the sensor, leveraging the errors caused by the discretization. However, this will equally affect the capability to represent dynamic objects in the environment at far distances, since several observations will be needed in order to change the occupancy state of such cells. 

Notice that changing the probabilities assigned to the sensor model $P_{occ}$ and $P_{free}$ will change the result of the reconstruction. Assigning a $P_{occ}$ considerably more influent than $P_{free}$ will increase the chances of representing partially occupied cells as completely occupied, but will affect the time of the approach to update dynamic cells in the map.

\section{Conclusions}

With this work we have introduced a method to update the occupancy probabilities of cells in a voxel grid in a more statistical way, by considering the traversed distances of all rays within the cells and applying a weighting function to the measurements according to the cell-sensor distance. Discretization inaccuracies are reduced by applying our second approach. These inaccuracies might be prevented by assigning a higher influence to the hit observations but this affects the capability of modeling dynamic environments.

Future work might include the use of synthetic data from simulators in order to quantitatively measure the improvement of our methods. 

\bibliographystyle{IEEEtran}
\bibliography{References}

\begin{thebibliography}{10}
\providecommand{\url}[1]{#1}
\csname url@samestyle\endcsname
\providecommand{\newblock}{\relax}
\providecommand{\bibinfo}[2]{#2}
\providecommand{\BIBentrySTDinterwordspacing}{\spaceskip=0pt\relax}
\providecommand{\BIBentryALTinterwordstretchfactor}{4}
\providecommand{\BIBentryALTinterwordspacing}{\spaceskip=\fontdimen2\font plus
\BIBentryALTinterwordstretchfactor\fontdimen3\font minus
  \fontdimen4\font\relax}
\providecommand{\BIBforeignlanguage}[2]{{%
\expandafter\ifx\csname l@#1\endcsname\relax
\typeout{** WARNING: IEEEtran.bst: No hyphenation pattern has been}%
\typeout{** loaded for the language `#1'. Using the pattern for}%
\typeout{** the default language instead.}%
\else
\language=\csname l@#1\endcsname
\fi
#2}}
\providecommand{\BIBdecl}{\relax}
\BIBdecl

\bibitem{Thrun02roboticmapping:}
S.~Thrun, ``Robotic mapping: A survey,'' in \emph{Exploring Artificial
  Intelligence in the New Millenium}.\hskip 1em plus 0.5em minus 0.4em\relax
  Morgan Kaufmann, 2002.

\bibitem{Curless1999}
B.~Curless and M.~Levoy, ``A volumetric method for building complex models from
  range images,'' vol.~3, 09 1999.

\bibitem{Newcombe6162880}
R.~A. Newcombe, S.~Izadi, O.~Hilliges, D.~Molyneaux, D.~Kim, A.~J. Davison,
  P.~Kohi, J.~Shotton, S.~Hodges, and A.~Fitzgibbon, ``Kinectfusion: Real-time
  dense surface mapping and tracking,'' in \emph{2011 10th IEEE International
  Symposium on Mixed and Augmented Reality}, Oct 2011, pp. 127--136.

\bibitem{Moravec-1985-15232}
H.~Moravec and A.~E. Elfes, ``High resolution maps from wide angle sonar,'' in
  \emph{Proceedings of the 1985 IEEE International Conference on Robotics and
  Automation}, March 1985, pp. 116 -- 121.

\bibitem{Thrun:2005:PR:1121596}
S.~Thrun, W.~Burgard, and D.~Fox, \emph{Probabilistic Robotics (Intelligent
  Robotics and Autonomous Agents)}.\hskip 1em plus 0.5em minus 0.4em\relax The
  MIT Press, 2005.

\bibitem{30717Bares}
J.~Bares, M.~Hebert, T.~Kanade, E.~Krotkov, T.~Mitchell, R.~Simmons, and
  W.~Whittaker, ``Ambler: an autonomous rover for planetary exploration,''
  \emph{Computer}, vol.~22, no.~6, pp. 18--26, June 1989.

\bibitem{100111Herbert}
M.~Herbert, C.~Caillas, E.~Krotkov, I.~S. Kweon, and T.~Kanade, ``Terrain
  mapping for a roving planetary explorer,'' in \emph{Proceedings, 1989
  International Conference on Robotics and Automation}, May 1989, pp. 997--1002
  vol.2.

\bibitem{Triebel4058725}
R.~Triebel, P.~Pfaff, and W.~Burgard, ``Multi-level surface maps for outdoor
  terrain mapping and loop closing,'' in \emph{2006 IEEE/RSJ International
  Conference on Intelligent Robots and Systems}, Oct 2006, pp. 2276--2282.

\bibitem{Milstein2008og}
A.~Milstein, ``Occupancy grid maps for localization and mapping,'' 06 2008.

\bibitem{DIA201713841}
R.~Dia, J.~Mottin, T.~Rakotovao, D.~Puschini, and S.~Lesecq, ``Evaluation of
  occupancy grid resolution through a novel approach for inverse sensor
  modeling,'' \emph{IFAC-PapersOnLine}, vol.~50, no.~1, pp. 13\,841 -- 13\,847,
  2017, 20th IFAC World Congress.

\bibitem{Hornung2013}
A.~Hornung, K.~M. Wurm, M.~Bennewitz, C.~Stachniss, and W.~Burgard, ``Octomap:
  an efficient probabilistic 3d mapping framework based on octrees,''
  \emph{Autonomous Robots}, vol.~34, no.~3, pp. 189--206, Apr 2013.

\bibitem{Elfes30720}
A.~Elfes, ``Using occupancy grids for mobile robot perception and navigation,''
  \emph{Computer}, vol.~22, no.~6, pp. 46--57, June 1989.

\bibitem{7857032Schaefer}
A.~Schaefer, L.~Luft, and W.~Burgard, ``An analytical lidar sensor model based
  on ray path information,'' \emph{IEEE Robotics and Automation Letters},
  vol.~2, no.~3, pp. 1405--1412, July 2017.

\bibitem{Geiger:2013:VMR:2528331.2528333}
A.~Geiger, P.~Lenz, C.~Stiller, and R.~Urtasun, ``Vision meets robotics: The
  kitti dataset,'' \emph{Int. J. Rob. Res.}, vol.~32, no.~11, pp. 1231--1237,
  Sep. 2013.

\end{thebibliography}

  \appendix

\subsection{Density Function $\rho(d)$ --- Development and Validation}\label{FirstAppendix}

This section explains the calculation of the density function mentioned in section \ref{Weight_funct}, that was used for our weighting function $w(d)$. This function has been obtained by making several approximations, but still models the reality accurately enough.

\begin{figure}[H]
	\begin{center}
		\includegraphics[width=0.38\textwidth]{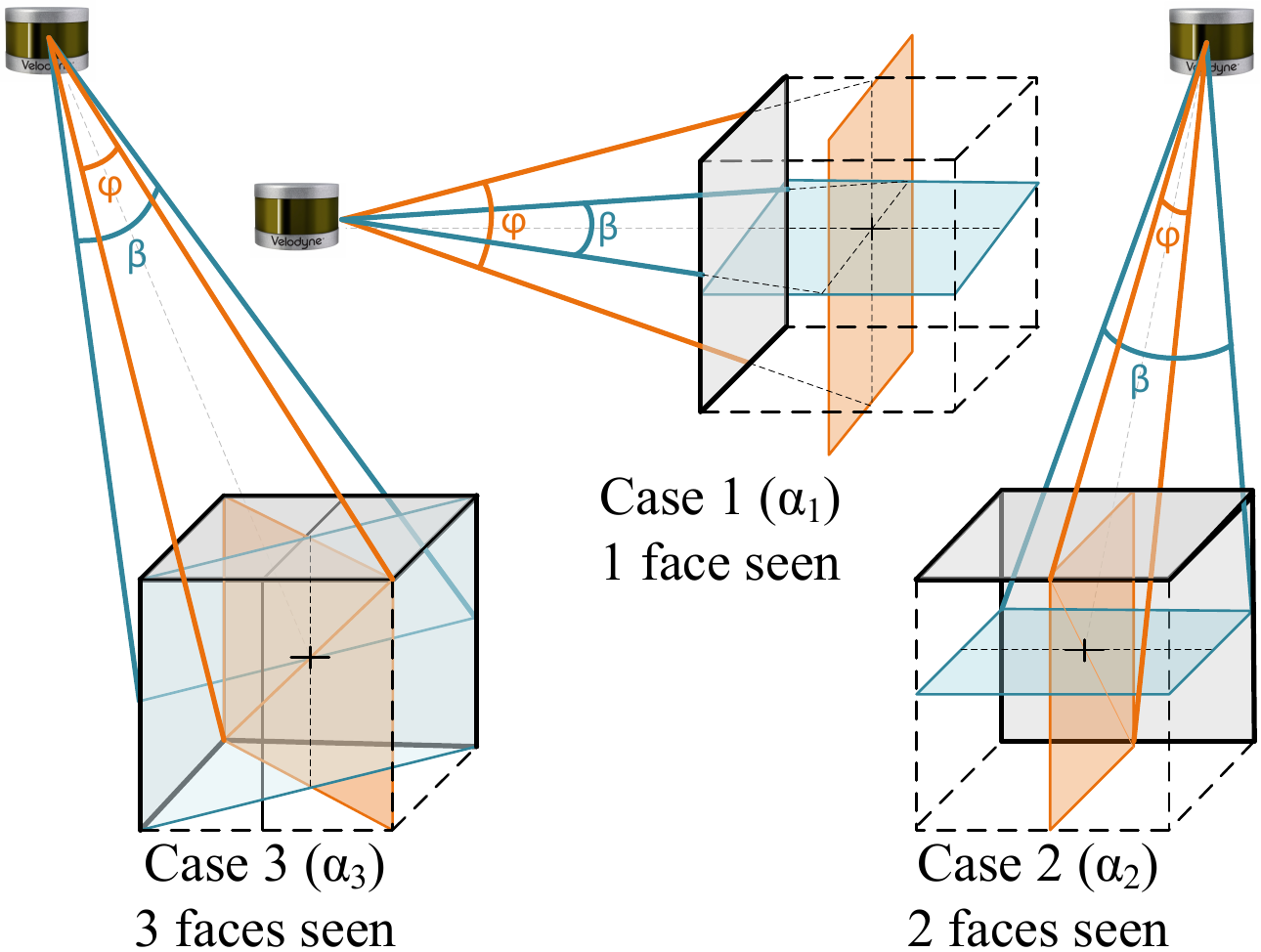}
		\caption{Cases considered in order to model the number of rays that traverse a voxel of size $\omega$ at distance $d$. Case 1: 3 faces seen by the sensor (top, left and rear). Case 2: 1 face seen by the sensor (left). Case 3: 2 faces seen by the sensor (top and rear).}
		\label{fig:WeightFun}
	\end{center}
\end{figure}

\vspace{-0.2cm}
To approximate the density of rays that can traverse a voxel $v$ of size $\omega$ at distance $d$, we consider three particular cases according to the number of faces of the voxel seen from the sensor (from one to three). This will depend on the relative position of the voxels with relation to the sensor in the grid. Referring to Fig. \ref{fig:WeightFun}, the three cases can be modeled as:

\small
\begin{subequations}
	\begingroup
	\thinmuskip=\muexpr\thinmuskip*5/8\relax
	\medmuskip=\muexpr\medmuskip*5/8\relax  
	\begin{equation}
	\alpha_1(d) = \frac{2}{\varphi_s} \tan^{-1}\left( \frac{\omega}{2d-\omega}\right)
	\frac{2}{\theta_s} \tan^{-1}\left( \frac{\omega}{2d-\omega}\right)
	\end{equation}  
	\begin{equation}
	\alpha_2(d) = \frac{2}{\varphi_s} \tan^{-1}\left( \frac{\sqrt{2}\omega}{2d-\sqrt{2}\omega}\right)
	\frac{2}{\theta_s} \tan^{-1}\left( \frac{\omega}{2d-\sqrt{2}\omega}\right)
	\end{equation}
	\begin{equation}
	\alpha_3(d) = \frac{2}{\varphi_s} \tan^{-1}\left( \frac{\sqrt{3}\omega}{2d-\sqrt{3}\omega}\right)
	\frac{2}{\theta_s} \tan^{-1}\left( \frac{\sqrt{2}\omega}{2d-\sqrt{3}\omega}\right)
	\end{equation}
\end{subequations}
\endgroup
\normalsize

\vspace{+0.3cm}

\noindent where $\varphi_s$ and $\theta_s$ correspond to the vertical and horizontal angular resolutions of the sensor. Notice that the particular cases considered simplify the calculations but do not brings an exact model, the position and orientation of the sensor and the voxel would have to be considered for this. Cases 3 and 1 represent upper and lower bounds, respectively. 

The modeling function $\rho(d)$ can be then considered as the weighted arithmetic mean of these three values (Eq. \ref{eq:ModellingFunction}). The weights are assigned considering the percentage of voxels that fit in any of the three different cases at the surface of a sphere of radius $d$.

\begin{equation}\label{eq:ModellingFunction}
\rho(d) = \frac{\eta_1(d) \, \alpha_1(d) + \eta_2(d) \, \alpha_2(d) + \eta_3(d) \, \alpha_3(d)}{\eta_t(d)}
\end{equation} 

\noindent where $\eta_t(d)$ approximates the number of voxels of size $\omega$ at the surface of the sphere:

\begin{equation}\label{eq:totalvoxels}
\eta_t(d) \approx \frac{4 \, \pi \, d^{2}}{\omega^{2}}
\end{equation} 

\begin{figure}[t]
	\centering
	\subfigure[]{\label{fig:validation_omega0_8}\includegraphics[width=40mm]{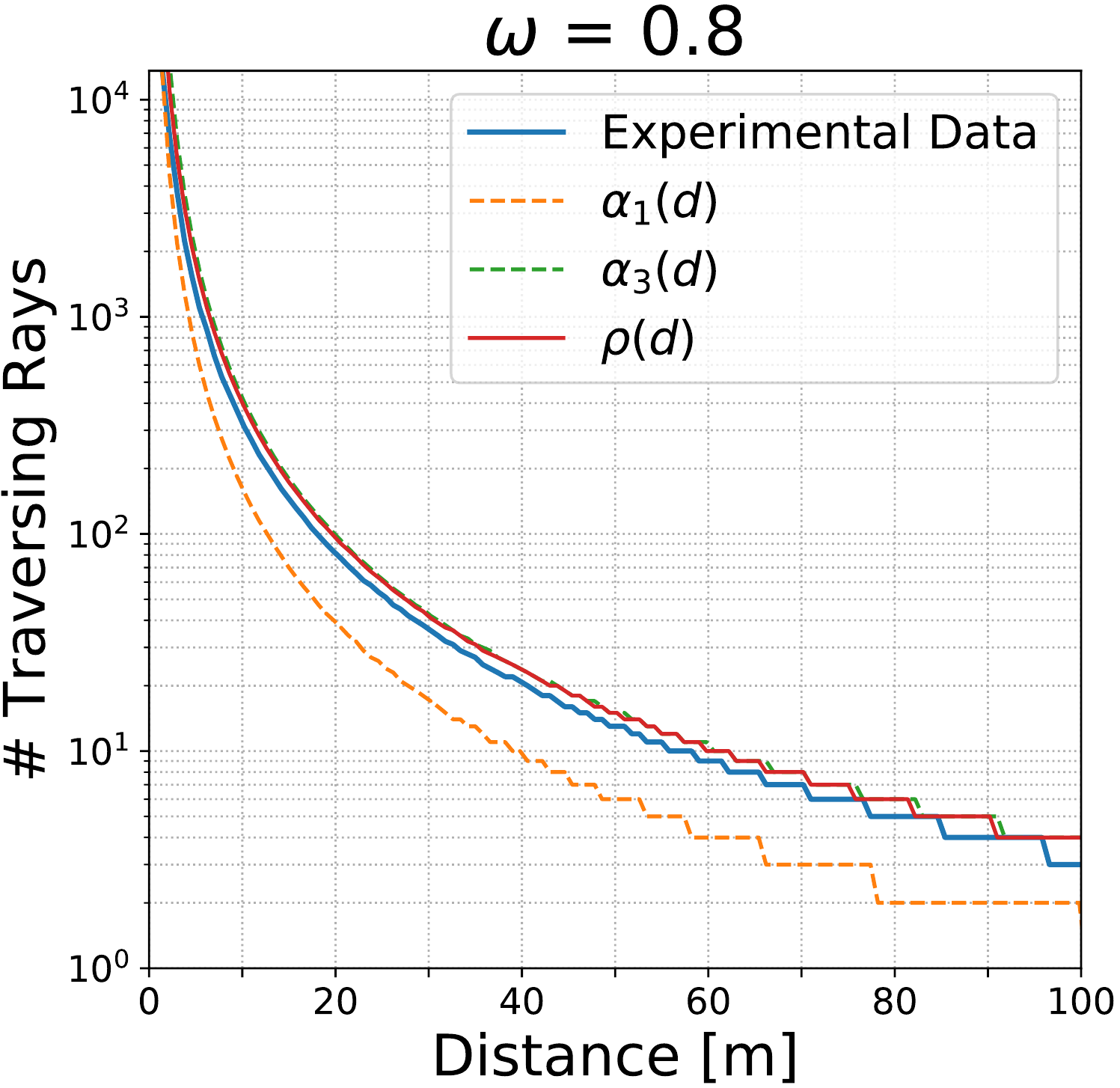}}
	\subfigure[]{\label{fig:validation_omega0_6}\includegraphics[width=40mm]{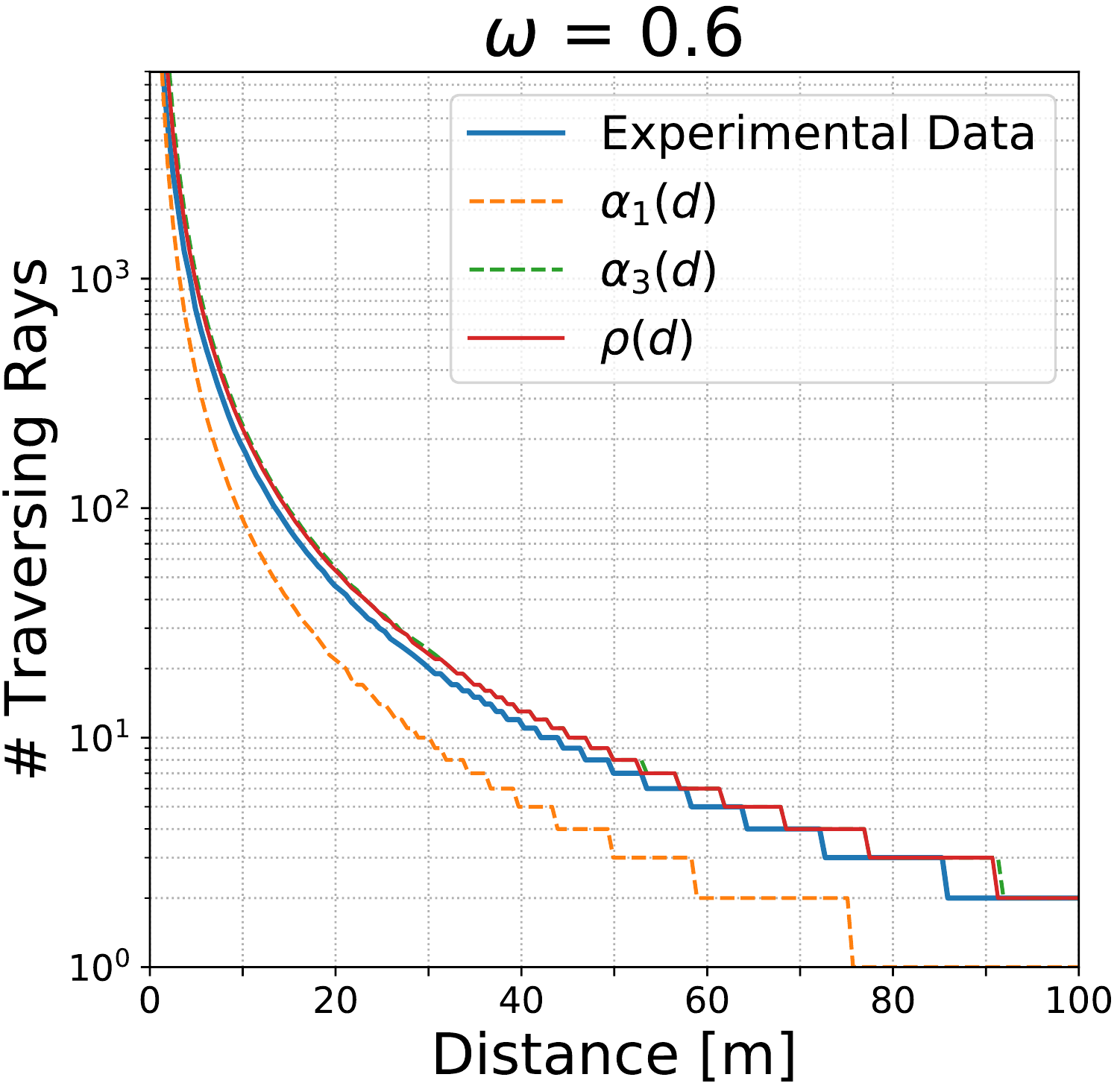}}
	\vspace{-0.1cm}
	\caption{Validation of the Modeling Function $\rho(d)$. Experimental data was compared with the model by inserting a spherical synthetic pointcloud of radius $r$ in a voxel grid of size $\omega$. (a) $\omega = 0.8$. (b) $\omega = 0.6$}
	\vspace{-0.3cm}
	\label{fig:ModelingFunValidation}
\end{figure}

If we consider the sensor placed at the center of the sphere, the number of voxels where two faces can be seen $\eta_2(d)$, are the ones placed within the perpendicular planes to the axes ($Oxy$, $Oxz$ and $Oyz$), except for the voxels aligned with the axes where only one face can be seen $\eta_1(d)$. The rest of voxels are represented by $\eta_3(d)$.

\begin{subequations}
	\begingroup
	\thinmuskip=\muexpr\thinmuskip*5/8\relax
	\medmuskip=\muexpr\medmuskip*5/8\relax  
	\begin{equation}
	\eta_1(d) = 6
	\end{equation}  
	\begin{equation}
	\eta_2(d) \approx 3\left(\frac{2 \, \pi \, d}{w}\right) - 12
	\end{equation}
	\begin{equation}
	\eta_3 \approx \frac{4 \, \pi \, d^{2}}{\omega^{2}} - \eta_1(d) - \eta_2(d)
	\end{equation}
\end{subequations}
\endgroup
\vspace{+0.05cm}

Notice that for larger distances, most of the voxels will belong to $\eta_3$. Moreover, if $d \gg w$, then $\eta_3$ can be approximated to $\eta_t$ and our modeling function can be simplified as $\rho(d) \approx \alpha_3(d)$.

In order to validate our model, we compare its results with the ones obtained by inserting a synthetic spherical pointcloud in a voxel grid at different resolutions and counting the number of rays traversing each cell in a $100m$ distance range, this can be seen on Fig. \ref{fig:ModelingFunValidation}. Notice that the green and orange curves represent the upper and lower bounds $\alpha_3(d)$ and $\alpha_1(d)$, respectively. The blue line represents the results obtained with the experiment, which are close to the upper bound since as it was stated previously, if $d \gg w$ most of the voxels are represented by $\alpha_3(d)$. The density function $\rho(d)$, as presented in Eq. \ref{eq:ModellingFunction} can also be seen in red. Once our modeling function has been validated, our weighting function is then derived from it as defined in Eq. \ref{eq:TheWeightingFunction}. 

\end{document}